\documentclass[conference]{IEEEtran}
\IEEEoverridecommandlockouts
\usepackage{cite}
\usepackage{amsmath,amssymb,amsfonts}
\usepackage{algorithmic}
\usepackage{graphicx}
\usepackage{textcomp}
\usepackage{xcolor}
\def\BibTeX{{\rm B\kern-.05em{\sc i\kern-.025em b}\kern-.08em
    T\kern-.1667em\lower.7ex\hbox{E}\kern-.125emX}}
\begin{document}

\title{ Ocular Disease Classification using 
Convolutional Neural Network with Deep Convolutional Generative
Adversarial Network\\
{\footnotesize \textsuperscript{}}
\thanks{}
}

\author{\IEEEauthorblockN{Arun Kunwar}
\IEEEauthorblockA{\textit{Institute of Engineering, Nepal} \\
arunjung1991@gmail.com}
\and

\IEEEauthorblockN{ *Dibakar Raj Pant}
\IEEEauthorblockA{\textit{Corresponding Author} \\
\textit{Institute of Engineering, Nepal}\\
drpant@ioe.edu.np}
\and

\IEEEauthorblockN{ Jukka Heikkonen}
\IEEEauthorblockA{\textit{University of Turku, Turku, Finland}\\
jukhei@utu.fi}
\and

\IEEEauthorblockN{ Rajeev Kanth}
\IEEEauthorblockA{\textit{Savonia University of Applied Sciences, Kuopio, Finland}\\
rajeev.kanth@savonia.fi}

}

\maketitle

\begin{abstract}
The Convolutional Neural Network (CNN) has shown impressive performance in image classification because of its strong learning capabilities. However, it demands a substantial and balanced dataset for effective training. Otherwise, networks frequently exhibit over fitting and struggle to generalize to new examples. Publicly available dataset of fundus images of ocular disease is insufficient to train any classification model to achieve satisfactory accuracy. So, we propose Generative Adversarial Network(GAN) based data generation technique to synthesize dataset for training CNN based classification model and later use original disease containing ocular images to test the model. During testing the model classification accuracy with the original ocular image, the model achieves an accuracy rate of 78.6\% for myopia, 88.6\% for glaucoma, and 84.6\% for cataract, with an overall classification accuracy of 84.6\%.
\end{abstract}


\section{Introduction}
Ocular diseases, including myopia, glaucoma, cataracts, diabetic retinopathy (DR), and age-related macular degeneration (AMD), present a significant threat to vision in the modern world\cite{b1}. These conditions can lead to permanent vision loss if not detected and treated promptly. The main challenge in managing these diseases is their often subtle or absent early symptoms, requiring the expertise of skilled medical professionals for accurate diagnosis. Early detection is paramount, as it offers the best chance for successful intervention and preserving vision. Late diagnosis can result in more extensive and costly treatments, with reduced prospects for restoring full vision\cite{b2}\cite{b3}. Diabetic retinopathy, a common complication of diabetes, underscores the importance of regular eye examinations for diabetic patients. Additionally, AMD stands out as a cause of irreversible vision loss in Western countries, often due to delays in diagnosis and treatment\cite{b4}. The integration of artificial intelligence technology has the potential to support primary ophthalmologists in their diagnostic processes by leveraging extensive medical data. This collaboration holds the promise of enhancing the accuracy and effectiveness of eye disease diagnosis and treatment within primary healthcare facilities\cite{b5}. \\ 
In recent times, there has been significant progress in the advancement of deep learning models within the realm of computer vision, leading to substantial improvements over conventional techniques. A noteworthy model in this domain is the convolutional neural network (CNN), which stands out due to its remarkable capacity for representing complex information, thereby compensating for the limitations of traditional feature extraction methods. A. Govindaiah et. al. \cite{6} studied the ability to effectively acquire intricate image features using (Visual Geometry Group)VGG-16 architecture, and they have demonstrated impressive performance in image classification of medical images.\\ 
In the context of ocular image analysis, the process of manually annotating these images necessitates the expertise of a medical professional, typically an ophthalmologist. Furthermore, due to privacy concerns, obtaining readily available and annotated medical images containing diseases can be a challenging endeavor. When it comes to training deep learning models for image classification tasks, it's essential to work with a sufficiently large dataset to prevent overfitting issues. This research paper explores two key approaches: firstly, the utilization of Deep Convolutional Generative Adversarial Networks as a method of data augmentation to expand the training dataset; and secondly, employing the synthetic images generated through DCGANs to train a Convolutional Neural Network model for the classification of diseases present in ocular images

\section{Related Work}

In recent years, the integration of Deep Convolutional Generative Adversarial Networks (DCGANs) into the domain of medical image classification has gained significant attention from researchers. This approach addresses a common challenge in this field - the scarcity of diverse and extensive labeled datasets. Previous studies have primarily focused on the detection of medical conditions using deep learning models trained on limited data. Generative Adversarial Networks have been employed for data augmentation with the goal of enhancing the training of Convolutional Neural Networks \cite{7}. DCGANs offer a compelling solution by generating synthetic medical images that closely mimic real ones. These synthetic images can help bridge the gap in data scarcity and class imbalance, thereby enhancing the robustness and accuracy of medical image classifiers.\\
Zeid Baker et. al.\cite{8} studied and employed Generative Adversarial Networks to generate synthetic images that closely resemble the real dataset, with the aim of expanding the available dataset. The research comprised two distinct experiments. In the first experiment, they fine-tuned a Deep Convolutional Generative Adversarial Network  specifically for a given dataset. The second experiment focused on assessing how the introduction of synthetic data impacted the accuracy of a classification task. The researchers conducted these experiments using three different datasets: MNIST, Fashion-MNIST, and Flower photos. Their findings suggested that the effectiveness of DCGAN in increasing model accuracy depended on the nature of the dataset and highlighted the significant role that data preprocessing played in the performance of DCGANs, a consideration applicable to most machine learning algorithms.\\ 
Wu, Qiufeng et. al. \cite{9}employed deep convolutional generative adversarial networks to augment their dataset with generated images, alongside original images, for the purpose of identifying Tomato. Their findings demonstrated that the images generated by DCGAN not only expanded the dataset size but also introduced diverse characteristics, ultimately leading to improved model generalization. They utilized a GoogLeNet classifier to train and test the model on five different classes of tomato leaf images. 

\section{Methodology}
\subsection{Block Diagram of the System}
\begin{figure}[htbp]
\centerline{\includegraphics[width=0.5\textwidth]{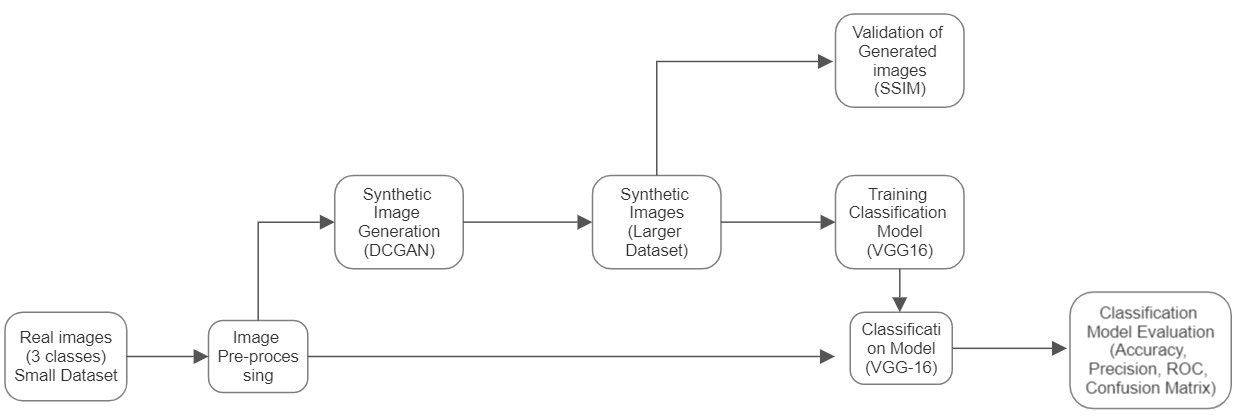}}
\caption{Block Diagram of the System}
\label{fig:block}
\end{figure}

Figure \ref{fig:block} depicts the overall functioning of the implemented system for the classification of ocular disease. Three different classes including glaucoma(G), cataract(C) and myopia (M) are included for the multi-class classification of disease of ocular image based on CNN to directly detect one ocular diseases in the retinal ocular images. As the available data set is not sufficient, the large data-set of ocular images is generated using DCGAN from available less data-set. Secondly, using the large data set, a CNN based model is developed to correctly classify the disease in the original ocular image.
\subsection{Dataset}
The dataset used in this study originates from the "International Competition on Ocular Disease Intelligent Recognition," which was sponsored by Peking University. This dataset comprises authentic patient data that was gathered by Shanggong Medical Technology Co. Ltd. from various hospitals and medical centers across China. The training dataset is a well-organized ophthalmic database containing information from 3,500 patients. It includes data such as patient age, color fundus photographs from both left and right eyes, as well as diagnostic keywords provided by doctors.There are eight different classes of diseases including normal(N), diabetic retinopathy(D), glaucoma(G), cataract(C), AMD(A), hypertensive retinopathy (H), myopia (M) and
other diseases/abnormalities (O). Morever,few ocular images in the dataset also contains two diseases making it multilabel classed image. However we have considered only single labelled ocular images for the study.

\subsection{Data Augmentation}
\begin{figure}[htbp]
\centerline{\includegraphics[width=0.5\textwidth]{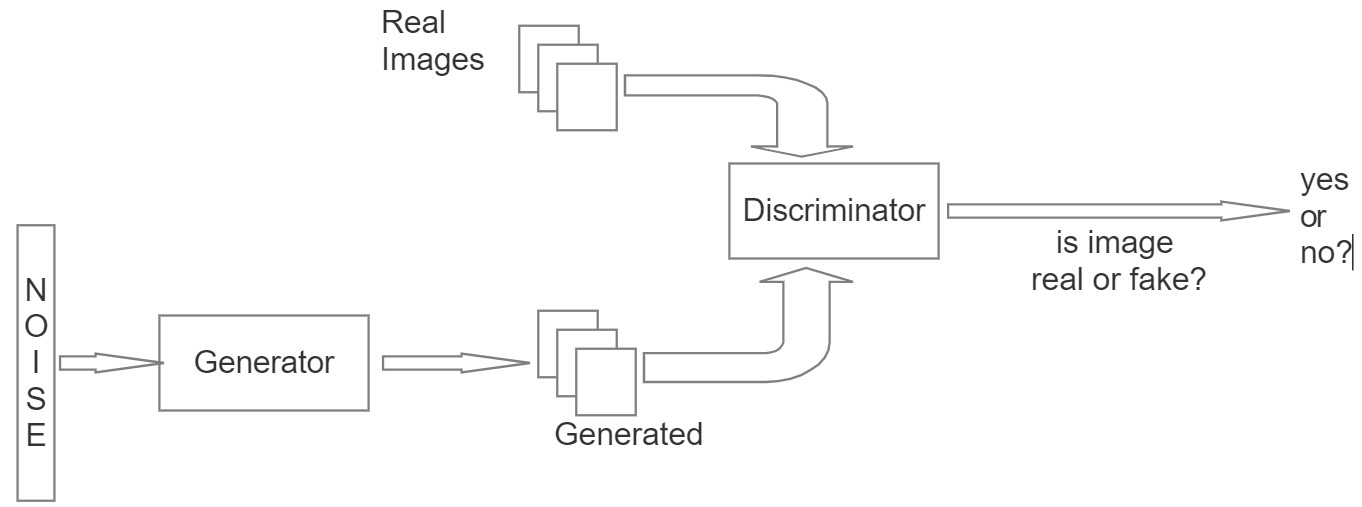}}
\caption{The structure of Generative Adversarial Network(GAN).}
\label{fig: GAN}
\end{figure}

As previously stated, datasets containing ocular images are notably limited in quantity and often exhibit imbalances. Consequently, DCGAN approach has been utilized to address this challenge. Generative Adversarial Networks generate images or data samples that closely mirror the feature distribution of the original dataset\cite{10}.\\
In the framework of a Generative Adversarial Network as shown in figure \ref{fig: GAN}, there are two key models, namely the generator and the discriminator, both of which are concurrently trained through an adversarial process. The discriminator's role is to acquire the ability to differentiate between real and counterfeit images, whereas the generator's objective is to produce images that closely resemble authentic ones. The training process continues until the discriminator reaches a point where it can no longer distinguish between genuine and synthetic images. GANs have demonstrated their effectiveness in various applications, including the generation of highly realistic human faces, medical image analysis, addressing class imbalance issues, and numerous other domains.

\begin{equation}
\min_G \max_D \mathbb{E}_{x \sim P_r}[\log(D(x))] + \mathbb{E}_{\tilde{x} \sim P_g}[\log(1 - D(\tilde{x}))]
\end{equation}

where:
\begin{itemize}
\item $\min_G$ represents the minimization with respect to the generator $G$.
\item $\max_D$ represents the maximization with respect to the discriminator $D$.
\item $\mathbb{E}_{x \sim P_r}$ denotes the expected value over real data samples $x$ drawn from the distribution $P_r$.
\item $\mathbb{E}_{\tilde{x} \sim P_g}$ denotes the expected value over generated data samples $\tilde{x}$ drawn from the generator's distribution $P_g$.
\item $D(x)$ represents the discriminator's output for a real data sample $x$.
\item $D(\tilde{x})$ represents the discriminator's output for a generated data sample $\tilde{x}$.
\end{itemize}

The objective function for the discriminator loss in a Generative Adversarial Network (GAN) is given by:

\begin{equation}
\mathcal{L}_D = -\mathbb{E}_{x \sim P_r}[\log(D(x))] - \mathbb{E}_{\tilde{x} \sim P_g}[\log(1 - D(\tilde{x}))]
\end{equation}

The objective function for the generator loss in a Generative Adversarial Network (GAN) is given by:

\begin{equation}
\mathcal{L}_G = -\mathbb{E}_{\tilde{x} \sim P_g}[\log(D(\tilde{x}))]
\end{equation}

\begin{figure}[h!]
\begin{center}
  	 \includegraphics[width=0.5\textwidth]{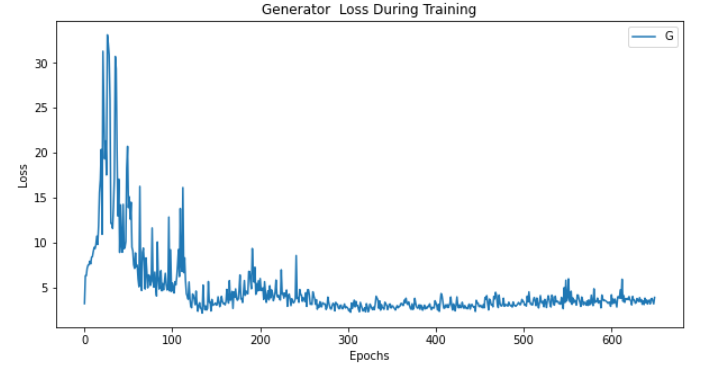}\\
  	 \caption{\label{fig:Generator Loss}Generator Loss }
\end{center}
\end{figure}

\begin{figure}[h!]
\begin{center}
  	 \includegraphics[width=0.5\textwidth]{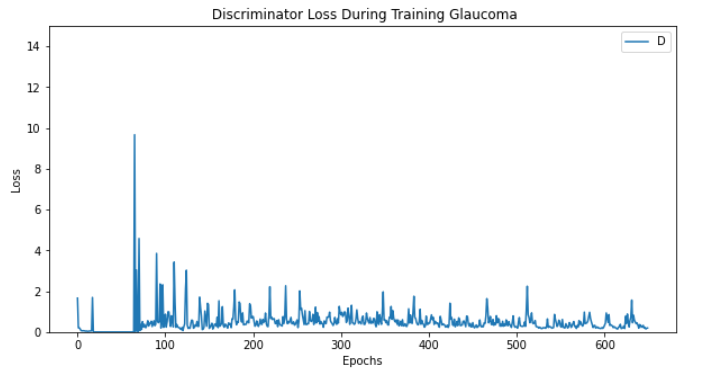}\\
  	 \caption{\label{fig:Discriminator Loss}Discriminator Loss }
\end{center}
\end{figure}

Goodfellow\cite{11} demonstrated that training GANs using the objective function described earlier, as referred to in Equations 1, 2, and 3 leads to issues of instability and lack of convergence.\\
In this research, we employed the Deep Convolutional Generative Adversarial Network, an extension of the Generative Adversarial Network. DCGAN is widely recognized as one of the most renowned and efficient GAN implementations, particularly suited for visual stimuli. The generator network is responsible for generating synthetic data samples, such as images, from random noise. It typically starts with a low-resolution noise vector and progressively up scales it through a series of transposed convolutional layers, gradually refining the generated output. The discriminator, on the other hand, acts as a binary classifier, distinguishing between real data samples and those produced by the generator. It is constructed using convolutional layers and pooling operations to process and analyze the input data, assigning a probability score to determine whether an input is real or synthetic. DCGANs are designed with specific architectural guidelines, such as the use of batch normalization and activation functions like ReLU, to ensure stable training and the generation of high-quality images. \\
DCGAN model with parameters shown in table  \ref{parasDCGAN} was built.
\begin{table}[htbp]
    \centering
     \caption{Parameters for DCGAN}
    \begin{tabular}{|c|c|c|}
        \hline
        \textbf{S.N.} & \textbf{Parameters} & \textbf{Values} \\
        \hline
         1. & Size of latent vector & 100 \\
        \hline
        2.& Number of channels & 3 \\
        \hline
        3. & Input Image Size & 64 X 64 X3\\
        \hline
        4. & Number of Epochs & 650\\
        \hline
        5. & Images per class & 300 approx\\
        \hline
        6. & Batch Size & 128\\
        \hline
        7. &  Learning Rate & 0.0002\\
        \hline
        8. & Optimizer & Adam\\
        \hline
    \end{tabular}
   
    \label{parasDCGAN}
\end{table}
\\
 Myopic, Glaucoma
and Cataract ocular images were the classes chosen for the generation. Total 850 images were available including all three classes of ocular images. With the above parameters set in the DCGAN model, Generator Model was built separately with three different class of images and 10,000 synthetic images for each classes of images were generated. Losses of Generator and Discriminator are shown in figure \ref{fig:Generator Loss}  and \ref{fig:Discriminator Loss} 

\begin{figure}[h!]
\begin{center}
  	 \includegraphics[width=0.45\textwidth]{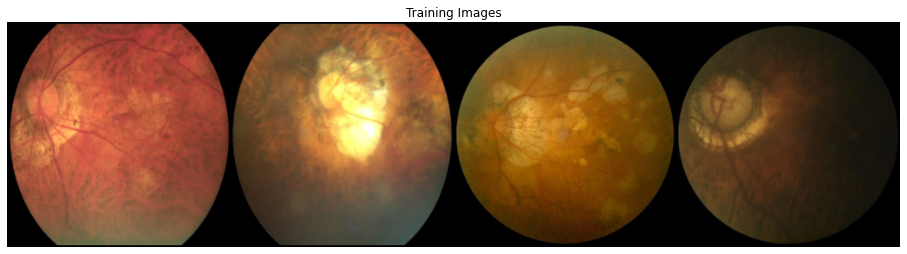}\\
  	 \caption{\label{fig:Real Myopic Images}Real Myopic Ocular Image Samples }
\end{center}
\end{figure}

\begin{figure}[h!]
\begin{center}
  	 \includegraphics[width=0.45\textwidth]{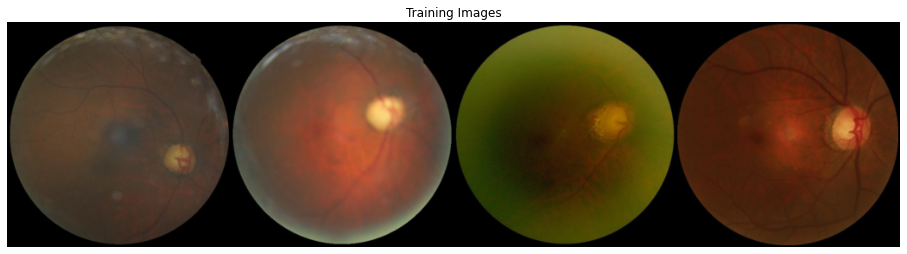}\\
  	 \caption{\label{fig:Real vs Fake Image Samples}Real Glaucoma Ocular Image Samples }
\end{center}
\end{figure}

\begin{figure}[h!]
\begin{center}
  	 \includegraphics[width=0.45\textwidth]{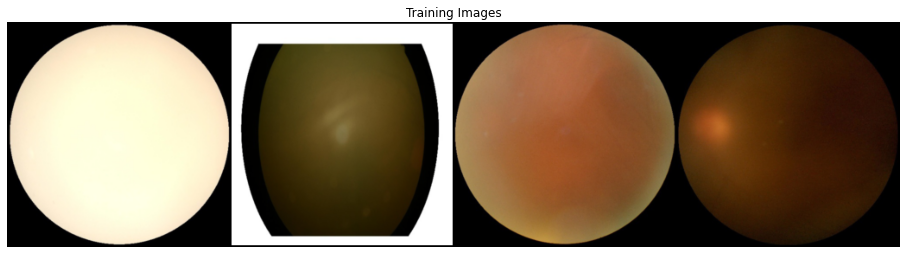}\\
  	 \caption{\label{fig:Real Cataract Fundus Images Samples Samples}Real Cataract Ocular Image Samples }
\end{center}
\end{figure}

Using the previously mentioned objective functions for both the generator and discriminator and loss function graph of generator and discriminator in equation 1, 2 and 3, synthetic images were generated with a resolution of 64x64 pixels and a latent vector size of 100.

\begin{figure}[h!]
\begin{center}
  	 \includegraphics[width=0.45\textwidth]{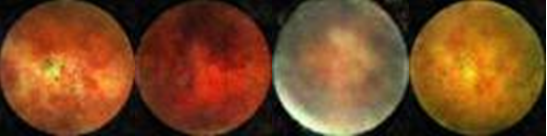}\\
  	 \caption{\label{fig:Generated Myopic Image Samples}Generated Myopic Ocular Image Samples }
\end{center}
\end{figure}

\begin{figure}[h!]
\begin{center}
  	 \includegraphics[width=0.45\textwidth]{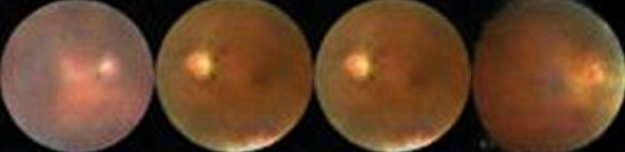}\\
  	 \caption{\label{fig:Generated Glaucoma Fundus Image Samples}Generated Glaucoma Ocular Image Samples }
\end{center}
\end{figure}

\begin{figure}[h!]
\begin{center}
  	 \includegraphics[width=0.45\textwidth]{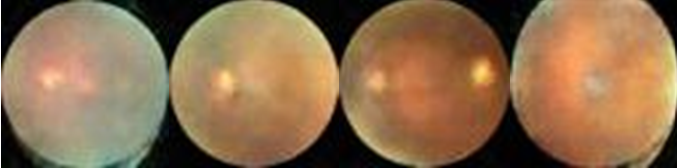}\\
  	 \caption{\label{figGenerated Cataract Fundus Image Samples}Generated Cataract Ocular Image Samples }
\end{center}
\end{figure}

\subsection{SSIM Test for Evaluation of Generated Images}
Chen et. al. \cite{11} studied the practical feasibility of structural Similarity Index to calculate the resemblance between real and synthetic images
For each classes 10,000 images was generated and SSIM test was done for the
generated images with the images of corresponding classes and the mean, maximum
and minimum SSIM value of each classes is mentioned in table II
\begin{table}[htbp]
     \caption{SSIM Value of Generated Images}
\begin{center}
\begin{tabular}{|c|c|c|c|}
         \hline
         \textbf{Class}  & \textbf{Max SSIM} & \textbf{Mean SSIM} & \textbf{Min SSIM} \\
        \hline
          Myopic &  0.80 & 0.66 & 0.43\\
        \hline
          Glaucoma   & 0.87 & 0.77 & 0.53\\
        \hline
         Cataract  & 0.90 & 0.76 & 0.48\\
        \hline
\end{tabular}
\label{SSIM}
\end{center}
\end{table}

\subsection{VGG-16 Classification Model}

 VGG-16\cite{13} comprises a total of 21 layers, consisting of 13 convolutional layers, 3 dense layers, and 5 max-pooling layers. However, only 16 of these layers have learnable parameters, hence the name VGG-16. The spatial dimension of the visual input stimuli processed by VGG is set at 224 × 224. Rectified Linear Unit (ReLU) serves as the activation function, with the Softmax classifier employed at the final layer.

\begin{figure}[h!]
\begin{center}
  	 \includegraphics[width=0.45\textwidth]{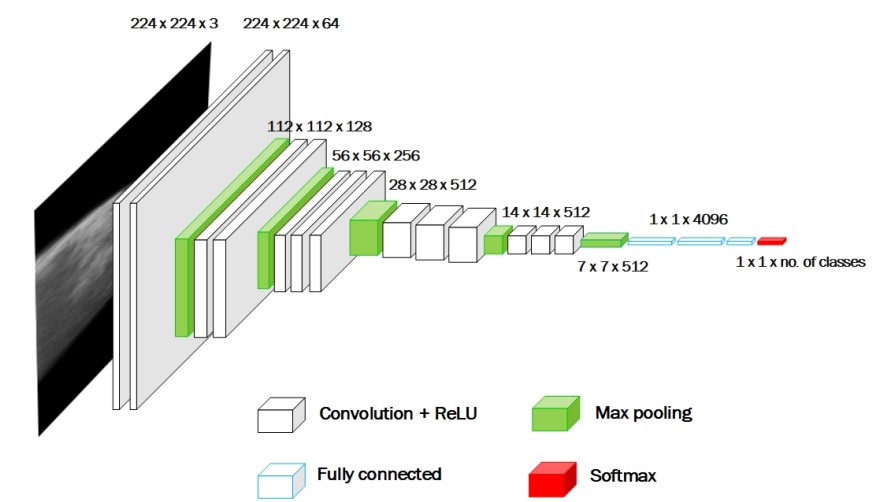}\\
  	 \caption{\label{fig:VGG-16} Architecture of VGG-16: Convolution, max-pooling, and dense layers}
\end{center}
\end{figure}

All three classes of ocular images generated by DCGAN are used as Training and Validation of the VGG-16 model. The training and validation of the VGG-16 model demonstrated stability and reliability when utilizing a larger and well-balanced generated dataset. The model was
trained and tested with following parameters in Table III.
\begin{table}[htbp]
    \centering
    \caption{Parameters for VGG-16}
    \begin{tabular}{|c|c|c|}
        \hline
        \textbf{S.N.} & \textbf{Parameters} & \textbf{Values} \\
        \hline
         1. & Number of Classes & 3 \\
        \hline
        2.& Number of channels & 3 \\
        \hline
        3. & Input Image Size & 224 X 224 X3\\
        \hline
        4. & Number of Epochs & 75\\
        \hline
        5. & Images per class & 10,000\\
        \hline
        6. & Batch Size & 32\\
        \hline
        7. &  Learning Rate & 0.001\\
        \hline
        8. & Optimizer & Adam\\
        \hline
    \end{tabular}
    
    \label{paralarge}
\end{table}
\section{Result and analysis}
\begin{figure}[h!]
\begin{center}
  	 \includegraphics[width=0.45\textwidth]{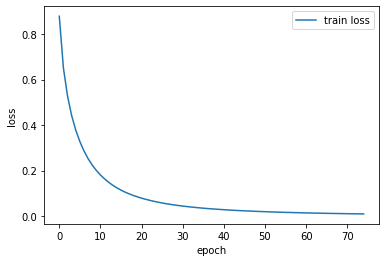}\\
  	 \caption{\label{fig:Training Loss}Training Loss }
\end{center}
\end{figure}
The stability of the model becomes evident when examining the accuracy and loss plots during both training and validation phases. These plots, which are based on a dataset consisting of 10,000 generated samples for each class, illustrate the consistent and reliable performance of the model.

\begin{figure}[h!]
\begin{center}
  	 \includegraphics[width=0.45\textwidth]{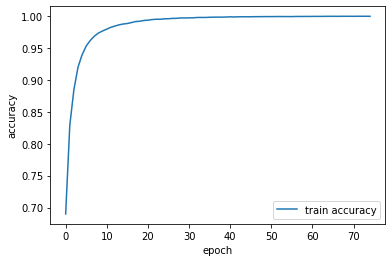}\\
  	 \caption{\label{fig:Training Accuracy}Training Accuracy }
\end{center}
\end{figure}

Training a model using DCGAN generated images alongside a pre-trained model like VGG-16 often yields improved training loss and accuracy for several reasons. Firstly, DCGAN-generated images are designed to closely mimic the distribution of real data, providing the network with a more diverse and representative training set. This augmentation reduces the risk of overfitting, as the model encounters a wider range of data patterns. Secondly, VGG-16 serves as a powerful feature extractor due to its depth and pre-trained weights on a large-scale dataset. It can capture intricate features from the DCGAN-generated images, enabling the model to learn richer representations and generalize better to real-world data. Additionally, the fine-tuning process aligns the VGG-16 model with the specific task, leveraging its hierarchical features for more effective classification. This combination of DCGAN-generated data and VGG-16's capabilities contributes to enhanced training loss convergence and accuracy, making it a beneficial strategy in various machine learning tasks, particularly in scenarios with limited real data.

\section{ROC and Confusion matrix}

\begin{table}[htbp]
    \centering
     \caption{Real and Generated Images }
    \begin{tabular}{|c|c|c|c|c|c|c|}
        \hline
        \textbf{S.N.} & \textbf{Class} & \textbf{Real Images}  & \textbf{Generated Images}  \\
        \hline
         1. & Myopic & 248 & 10,000\\
        \hline
         2. & Glaucoma & 361  & 10,000\\
        \hline
         3. & Cataract & 241 & 10,000\\
        \hline
        
    \end{tabular}
   
    \label{image}
\end{table}

When the VGG-16 classification model, which was initially trained using ocular images generated by a DCGAN, was evaluated on the original ocular images, it achieved a commendable accuracy of 84.6\% in distinguishing between three distinct classes of ocular images.
The ROC curve and the confusion matrix provide clear visual representations that effectively convey the model's accuracy.
\begin{figure}[h!]
\begin{center}
  	 \includegraphics[width=0.45\textwidth]{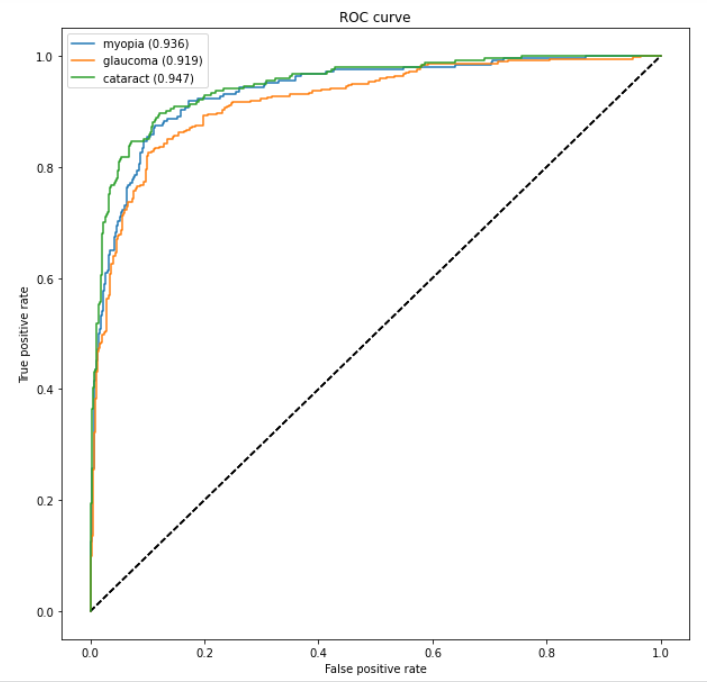}\\
  	 \caption{\label{fig:ROC Curve}ROC Curve}
\end{center}
\end{figure}

\begin{figure}[h!]
\begin{center}
  	 \includegraphics[width=0.45\textwidth]{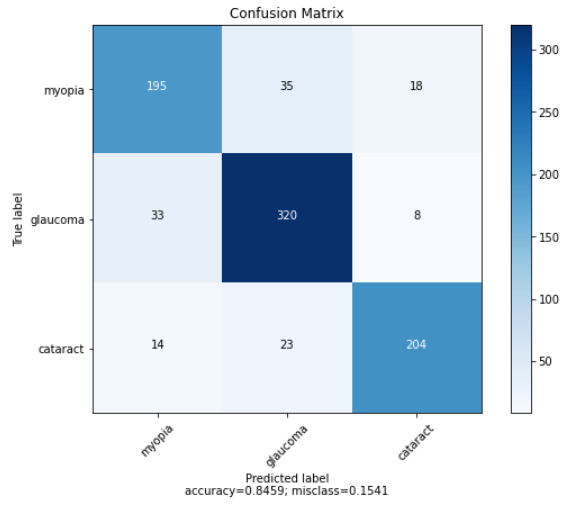}\\
  	 \caption{\label{fig:Confusion Matrix}Confusion Matrix  }
\end{center}
\end{figure}

\section{Discussion and Conclusion}

In our study, we showcase the utilization of a DCGAN-generated augmented dataset for constructing a classification model. Our results strongly suggest that employing GAN-based data augmentation is an effective approach for addressing the challenges posed by imbalanced and limited medical datasets. We provide evidence of improved AUC performance and present the results of the confusion matrix to support our findings. DCGAN is a robust method for generating high-quality medical images and expanding the training dataset. Additionally, utilizing synthetic image data in conjunction with CNNs can lead to enhanced accuracy in performance.\\
The model's 14\% misclassification rate could be attributed to a limited quantity of original ocular disease images available for training. Additionally, the DCGAN-generated images, which were at a resolution of 64x64 pixels, may have struggled to capture intricate disease-related details or features. Consequently, this limited the VGG-16 model's ability to reach its full potential during training. We noticed that the images we have obtained from DCGAN had some noise. Therefore, in the next
paper we will use denoising Autoencoder to enhance these images to improve the model accuracy.

\end{document}